\documentclass[11pt,a4paper]{article}
\usepackage[hyperref]{emnlp-ijcnlp-2019}
\usepackage{times}
\usepackage{latexsym}
\usepackage{amsmath}
\usepackage{algorithmic}
\usepackage{algorithm}
\usepackage{tabularx}
\usepackage{bm}
\usepackage{graphicx}
\usepackage{booktabs}
\usepackage{amsfonts}
\usepackage{multirow}
\usepackage{xcolor}
\usepackage{url}
\usepackage{commath}
\usepackage{boxedminipage}

\definecolor{gred}{RGB}{219,68,55}
\definecolor{gblue}{RGB}{66,133,244}
\definecolor{gyellow}{RGB}{244,180,0}
\definecolor{ggreen}{RGB}{15,157,88}
\definecolor{ggrey}{RGB}{115,115,115}

\aclfinalcopy 

\title{Select and Attend: Towards Controllable Content Selection in Text Generation}

\author{Xiaoyu Shen$^{1,2}$\thanks{{\hspace{1.5 mm}}Work mostly done while at RIKEN AIP. Correspondence to \tt{xshen@mpi-inf.mpg.de}}, 
Jun Suzuki$^{3,4}$, Kentaro Inui$^{3,4}$, Hui Su$^5$\\
\textbf{Dietrich Klakow$^{1}$ and Satoshi Sekine$^{4}$}\\
$^1$Spoken Language Systems (LSV), Saarland University, Germany\\
$^2$Max Planck Institute for Informatics, Saarland Informatics Campus, Germany\\
$^3$Tohoku University \hspace{5 mm} $^4$RIKEN AIP, Japan\\
$^5$Pattern Recognition Center, Wechat AI, Tencent Inc, China}

\date{}

\begin{document}
\maketitle
\begin{abstract}
Many text generation tasks naturally contain two steps: content selection and surface realization. Current neural encoder-decoder models conflate both steps into a black-box architecture. As a result, the content to be described in the text cannot be explicitly controlled. This paper tackles this problem by decoupling content selection from the decoder. The decoupled content selection is human interpretable, whose value can be manually manipulated to control the content of generated text. The model can be trained end-to-end without human annotations by maximizing a lower bound of the marginal likelihood. We further propose an effective way to trade-off between performance and controllability with a single adjustable hyperparameter. In both data-to-text and headline generation tasks, our model achieves promising results, paving the way for controllable content selection in text generation.\footnote{The source code is available on \url{https://github.com/chin-gyou/controllable-selection} }
\end{abstract}

\section{Introduction}
\label{sec: intro}
\begin{table}
\setlength{\columnwidth}{4pt}
\center
\footnotesize
\begin{tabular}{@{}l}
{ \parbox{7.5cm}{{\bf Source Sentence: }The \textcolor{ggreen}{sri lankan} government on \textcolor{gblue}{Wednesday} \textcolor{orange}{announced} the \textcolor{ggreen}{closure} of government \textcolor{ggreen}{schools} with immediate effect as a \textcolor{red}{military campaign} against tamil separatists escalated in the {north} of the country.}}\\\midrule

{ \parbox{7.5cm}{{\bf Selected : } \textcolor{ggreen}{sri lankan}, \textcolor{ggreen}{closure}, \textcolor{ggreen}{schools}}}\\
{ \parbox{7.5cm}{{\bf Text: } \textcolor{ggreen}{sri lanka closes schools} .}}\\\midrule
{ \parbox{7.5cm}{{\bf Selected : } \textcolor{ggreen}{sri lankan}, \textcolor{gblue}{Wednesday}, \textcolor{ggreen}{closure}, \textcolor{ggreen}{schools}}}\\
{ \parbox{7.5cm}{{\bf Text: } \textcolor{ggreen}{sri lanka closes schools} on \textcolor{gblue}{Wednesday}.}}\\\midrule
{ \parbox{7.5cm}{{\bf Selected : } \textcolor{ggreen}{sri lankan}, \textcolor{ggreen}{closure}, \textcolor{ggreen}{schools}, \textcolor{red}{military campaign}}}\\
{ \parbox{7.5cm}{{\bf Text: } \textcolor{ggreen}{sri lanka shuts down schools} amid \textcolor{red}{war} fears.}}\\\midrule
{ \parbox{7.5cm}{{\bf Selected : } \textcolor{ggreen}{sri lankan}, \textcolor{orange}{announced}, \textcolor{ggreen}{closure}, \textcolor{ggreen}{schools}}}\\
{ \parbox{7.5cm}{{\bf Text: } \textcolor{ggreen}{sri lanka} \textcolor{orange}{declares} \textcolor{ggreen}{closure} of \textcolor{ggreen}{ schools}.}}\\\midrule

\end{tabular}
\caption{\label{tab: intro}Headline generation examples from our model. We can generate text describing various contents by sampling different content selections. The selected source word and its corresponding realizations in the text are highlighted with the same color.} 
\end{table}
Many text generation tasks, e.g., data-to-text, summarization and image captioning, can be naturally divided into two steps: content selection and surface realization.  The generations are supposed to have two levels of diversity: (1) content-level diversity reflecting multiple possibilities of content selection (what to say) and (2) surface-level diversity reflecting the linguistic variations of verbalizing the selected contents (how to say)~\cite{reiter2000building,nema2017diversity}. Recent neural network models handle these tasks with the encoder-decoder (Enc-Dec) framework~\cite{sutskever2014sequence,bahdanau2015neural}, which simultaneously performs selecting and verbalizing in a black-box way. Therefore, both levels of diversity are entangled within the generation. This entanglement, however, sacrifices the controllability and interpretability, making it diffifcult to specify the content to be conveyed in the generated text ~\cite{qin2018learning,wiseman2018learning}.

With this in mind, this paper proposes decoupling content selection from the Enc-Dec framework to allow finer-grained control over the generation. Table~\ref{tab: intro} shows an example. We can easily modify the content selection to generate text with various focuses, or sample multiple paraphrases by fixing the content selection.

Though there has been much work dealing with content selection for the Enc-Dec, none of them is able to address the above concerns properly. Current methods can be categorized into the following three classes and have different limits:
\begin{enumerate}
    \item \textbf{Bottom-up}: Train a separate content selector to constrain the attention to source tokens~\cite{gehrmann2018bottom}, but the separate training of selector/generator might lead to discrepancy when integrating them together.
    \item \textbf{Soft-select}: Learn a soft mask to filter useless information~\cite{mei2016talk,zhou2017selective}. However, the mask is \emph{deterministic} without any probabilistic variations, making it hard to model the content-level diversity.
    \item \textbf{Reinforce-select}: Train the selector with reinforcement learning~\cite{chen2018fast}, which has high training variance and low diversity on content selection.
\end{enumerate}
In this paper, we treat the content selection as latent variables and train with amortized variational inference~\cite{kingma2013auto, mnih2014neural}. This provides a lower training variance than Reinforce-select. The selector and generator are co-trained within the same objective, the generations are thus more faithful to the selected contents than Bottom-up methods.
Our model is task-agnostic, end-to-end trainable and can be seamlessly inserted into any encoder-decoder architecture. On both the data-to-text and headline generation task, we show our model outperforms others regarding content-level diversity and controllability while maintaining comparable performance. The performance/controllability trade-off can be effectively adjusted by adjusting a single hyperparameter in the training stage, which constrains an upper bound of the conditional mutual information (CMI) between the selector and generated text~\cite{pmlr-v80-alemi18a,zhao2018information}. A higher CMI leads to stronger controllability with a bit more risk of text disfluency. 

In summary, our contributions are \textbf{(1)} systematically studying the problem of controllable content selection for Enc-Dec text generation, \textbf{(2)} proposing a task-agnostic training framework achieving promising results and \textbf{(3)} introducing an effective way to achieve the trade-off between performance and controllability.

\section{Background and Notation}
Let $X,Y$ denote a source-target pair. $X$ is a sequence of $x_1,x_2,\ldots,x_n$ and can be either some structured data or unstructured text/image depending on the task. $Y$ corresponds to $y_1,y_2,\ldots,y_m$ which is a text description of $X$. The goal of text generation is to learn a distribution $p(Y|X)$ to automatically generate proper text.

The Enc-Dec architecture handles this task with an encode-attend-decode process~\cite{bahdanau2015neural,xu2015show}. The encoder first encodes each $x_i$ into a vector $h_i$. At each time step, the decoder pays attentions to some source embeddings and outputs the probability of the next token by $p(y_t|y_{1:t-1},C_t)$. $C_t$ is a weighted average of source embeddings:
\begin{equation}
\label{eq: attention}
\begin{split}
C_t &= \sum_{i}\alpha_{t,i}h_{i}\\
\alpha_{t,i} &= \frac{e^{f(h_{i}, d_t)}}{\sum_j e^{f(h_{j}, d_t)}}
\end{split}
\end{equation}
$d_{t}$ is the hidden state of the decoder at time step $t$. $f$ is a score function to compute the similarity between  $h_i$ and $d_t$~\cite{luong2015effective}.

\section{Content Selection}
\begin{figure}[ht]
\centering
\centerline{\includegraphics[width=\columnwidth]{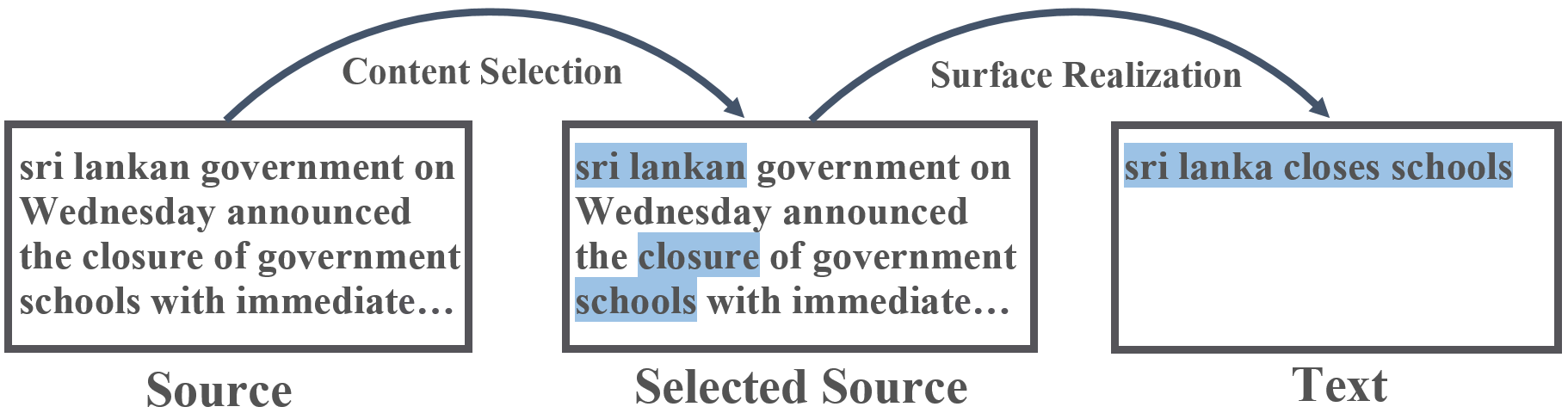}}
\caption{Model will select contents based on $\mathbf{B}(\gamma)$, then decode with $p_\theta(Y|X,\beta)$. Source-text pairs are available for training, but the ground-truth content selection for each pair is unknown.}
\label{fig: structure}
\end{figure}

Our goal is to decouple the content selection from the decoder by introducing an extra content selector. We hope the content-level diversity can be fully captured by the content selector for a more interpretable and controllable generation process. Following \citet{gehrmann2018bottom,yu2018operation}, we define content selection as a sequence labeling task. Let $\beta_1,\beta_2,\ldots,\beta_n$ denote a sequence of binary selection masks. $\beta_i=1$ if $h_i$ is selected and 0 otherwise. $\beta_i$ is assumed to be independent from each other and is sampled from a bernoulli distribution $\mathbf{B}(\gamma_i)$\footnote{\citet{devlin2018bert} have shown that excellent performance can be obtained by assuming such conditionally independence given a sufficiently expressive representation of $x$, though modelling a richer inter-label dependency is for sure beneficial~\cite{lei2016rationalizing, nallapati2017summarunner}.}. $\gamma_i$ is the bernoulli parameter, which we estimate using a two-layer feedforward network on top of the source encoder. Text are generated by first sampling $\beta$ from $\mathbf{B}(\gamma)$ to decide which content to cover, then decode with the conditional distribution $p_\theta(Y|X,\beta)$. The text is expected to faithfully convey all selected contents and drop unselected ones. Fig.~\ref{fig: structure} depicts this generation process. Note that the selection is based on the token-level \emph{context-aware} embeddings $h$ and will maintain information from the surrounding contexts. It encourages the decoder to stay faithful to the original information instead of simply fabricating random sentences by connecting the selected tokens.
For each source-target pair, the ground-truth selection mask is unknown, so training is challenging. In the following session, we discuss several training possibilities and introduce the proposed model in detail.

\subsection{Bottom-up}
The most intuitive way is training the content selector to target some heuristically extracted contents. For example, we can train the selector to select overlapped words between the source and target~\cite{gehrmann2018bottom}, sentences with higher tf-idf scores~\cite{li2018improving} or identified image objects that appear in the caption~\cite{wang2017diverse}. A standard encoder-decoder model is independently trained. In the testing stage, the prediction of the content selector is used to hard-mask the attention vector to guide the text generation in a bottom-up way. Though easy to train, Bottom-up generation has the following two problems: \textbf{(1)} The heuristically extracted contents might be coarse and cannot reflect the variety of human languages and \textbf{(2)} The selector and decoder are independently trained towards different objectives thus might not adapt to each other well.

\textbf{$\bm{\beta}$ as Latent Variable:} Another way is to treat $\beta$ as a latent variable and co-train selector and generator by maximizing the marginal data likelihood. By doing so, the selector has the potential to automatically explore optimal selecting strategies best fit for the corresponding generator component.

With this in mind. We design $p_\theta(Y|X,\beta)$ by changing the original decoder in the following way: (1) We initialize hidden states of the decoder from a mean pooling over selected contents to inform the decoder which contents to cover and (2) Unselected contents will be prohibited from being attended to:
\begin{align}
\label{eq: mask-attention}
\begin{split}
    &d_0 = \text{MLP}\left(\frac{1}{n}\left(\sum_{i}^{n}\beta_i h_i\right)\right)\\
&\alpha_{t,i} = \frac{e^{f\left(h_{i}, d_t\right)}\beta_i}{\sum_j e^{f\left(h_{j}, d_t\right)}\beta_j}
\end{split}
\end{align}
$d_0$ is the initial decoder hidden state and MLP denotes multi-layer-perceptron. 

Since computing the exact marginal likelihood $\log \mathbb{E}_{\beta \sim \mathbf{B}(\gamma)} p_\theta(Y|X,\beta)$ requires enumerating over all possible combinations of $\beta$ (complexity $\mathbf{O}(2^n)$), we need some way to efficiently estimate the likelihood.

\subsection{Soft-Select}
\label{sec: soft-select}
Soft-select falls back on a \emph{deterministic} network to output the likelihood function's first-order Taylor series approximation expanded at $\mathbb{E}_{\beta \sim \mathbf{B}(\gamma)} \beta$:
\begin{equation*}
\begin{split}
&\log \mathbb{E}_{\beta \sim \mathbf{B}(\gamma)} p_\theta(Y|X,\beta)\\\approx & \log [p_\theta(Y|X, \gamma)+\mathbb{E}_{\beta\sim \mathbf{B}(\gamma)}(\beta-\gamma)p'_\theta(Y|X, \gamma)]\\
=&\log p_\theta(Y|X, \gamma)
\end{split}
\end{equation*}
By moving the expectation into the decoding function, we can deterministically compute the likelihood by setting $\beta_i=\gamma_i$, reducing complexity to $\mathbf{O}(1)$. Each attention weight will first be ``soft-masked" by $\gamma$ before being passed to the decoder. soft-select is fully differentiable and can be easily trained by gradient descent. However, this soft-approximation is normally inaccurate, especially when $\mathbf{B}(\gamma)$ has a high entropy, which is common in one-to-many text generation tasks. The gap between $\log \mathbb{E}_{\beta \sim \mathbf{B}(\gamma)} p_\theta(Y|X,\beta)$ and $\log p_\theta(Y|X, \mathbb{E}_{\beta\sim \mathbf{B}(\gamma)})$ will be large~\cite{ma2017dropout,deng2018latent}. In practice, this would lead to unrealistic generations when sampling $\beta$ from the deterministically trained distribution.
\subsection{Reinforce-Select}
\label{sec: reinforced-select}
 Reinforce-select (RS)~\cite{ling2017coarse,chen2018fast} utilizes reinforcement learning to approximate the marginal likelihood. Specifically, it is trained to maximize a lower bound of the likelihood by applying the Jensen inequalily:
\begin{equation*}
\label{eq: hard-mask}
\log \mathbb{E}_{\beta \sim \mathbf{B}(\gamma)} p_\theta(Y|X,\beta)
\geq\mathbb{E}_{\beta \sim \mathbf{B}(\gamma)}\log p_\theta(Y|X,\beta)
\end{equation*}

The gradient to $\gamma$ is approximated with Monte-Carlo sampling by applying the REINFORCE algorithm~\cite{williams1992simple,glynn1990likelilood}. To speed up convergence, we pre-train the selector by some distant supervision, which is a common practice in reinforcement learning. REINFORCE is unbiased but has a high variance. Many research have proposed sophisticated techniques for variance reduction~\cite{mnih2014neural,tucker2017rebar,grathwohl2018backpropagation}. In text generation, the high-variance problem is aggravated because there exists multiple valid selections. Accurately estimating the likelihood becomes difficult. Another issue is its tendency to avoid stochasticity~\cite{raiko2014techniques}, which we will show in Sec~\ref{sec: results} that it results in low content-level diversity. 
\subsection{Variational Reinforce-Select}
 We propose Variational Reinforce-Select (VRS) which applies variational inference~\cite{kingma2013auto} for variance reduction. Instead of directly integrating over $\mathbf{B}(\gamma)$, it imposes a proposal distribution $q_\phi$ for importance sampling. The marginal likelihood is lower bounded by:
\begin{equation}
\label{eq: vhard-mask}
\begin{split}
&\log \mathbb{E}_{\beta \sim \mathbf{B}(\gamma)} p_\theta(Y|X,\beta)\\
&=\log \mathbb{E}_{\beta \sim q_\phi} \frac{p_\theta(Y,\beta|X)}{q_\phi(\beta)}\\
&\geq\mathbb{E}_{\beta \sim q_\phi}\log \frac{p_\theta(Y,\beta|X)}{q_\phi(\beta)}\\
&=\mathbb{E}_{\beta \sim q_\phi}\log p_\theta(Y|X,\beta) - KL(q_\phi||\mathbf{B}(\gamma))
\end{split}
\end{equation}
By choosing a proper $q_\phi$, the bound will be improved and the variance can be largely reduced compared with REINFORCE. If $q_\phi$ equals the posterior distribution $p_\theta(\beta|X,Y)$, the bound is tight and the variance would be zero~\cite{mnih2016variational}. We define $q_\phi(\beta|X,Y)$ as a mean-field distribution parameterized by a set of global parameters $\phi$ to approach the true posterior distribution. $\phi$, $\theta$ and $\gamma$ are simultaneously trained by minimizing the last tine of Eq.~\ref{eq: vhard-mask}. $q_\phi(\beta|X,Y)$ also allows us to further perform posterior inference: Given an arbitrary text $Y$ for a source $X$, we can infer which source contents are included in $Y$ (An example is given in Appendix~\ref{app: post}). 

In Eq.\ref{eq: vhard-mask},  the KL divergence term can be computed analytically. As for the independence assumption, it can be summed over each individual $\beta_i$. 
The likelihood term is differentiable to $\theta$ but not to $\phi$, we estimate the gradient to $\phi$ in Eq~\ref{eq: vhard-mask} by applying the REINFORCE estimator:
\begin{equation*}
\label{eq: vhard-train}
\begin{gathered}
\nabla_\phi\mathbb{E}_{\beta \sim q_\phi}\log p_\theta(Y|X,\beta)=\\
\mathbb{E}_{\beta \sim q_\phi} \nabla_\phi \log q_\phi(\beta|X,Y)(\log p_\theta(Y|X,\beta) - B)
\end{gathered}
\end{equation*}
$B$ is the control variate~\cite{williams1992simple}. The optimal $B$ would be~\cite{weaver2001optimal}:
\begin{equation*}
\label{eq: op-baseline}
B^{\ast} = \mathbb{E}_{\beta \sim q_\phi}\log p_\theta(Y|X,\beta)
\end{equation*}
which we set as a soft-select approximation:
\begin{equation*}
\label{eq: baseline}
B=\log p_\theta(Y|X,\mathbb{E}_{\beta \sim q_\phi} \beta)
\end{equation*}
 We estimate Eq.~\ref{eq: vhard-train} with a single sample from $q_\phi$ for efficiency. Though multiple-sample could potentially further tighten the bound and reduce the variance~\cite{burda2015importance,lawson2018learning,tucker2019doubly}, it brings significant computational overhead, especially in text generation tasks where the whole sentence needs to be decoded.
\subsection{Degree of Controllability}
\label{sec: trade-off}
In practice, when treating content selection as latent variables, the model tends to end up with a trivial solution of always selecting all source tokens~\cite{shen2018reinforced,pmlr-v80-ke18a}.
This behavior is understandable since Eq.~\ref{eq: mask-attention} strictly masks unselected tokens. Wrongly unselecting one token will largely deteriorate the likelihood. Under the maximum likelihood (MLE) objective, this high risk pushes the selector to take a conservative strategy of always keeping all tokens, then the whole model degenerates to the standard Enc-Dec and the selection mask loses effects on the generation. Usually people apply a penalty term to the selecting ratio when optimizing the likelihood:
\begin{equation}
    \label{eq: hh-alpha}
    \mathcal{L} + \lambda \abs{(\Bar{\gamma}-\alpha)}
\end{equation}
 $\mathcal{L}$ is the MLE loss function, $\Bar{\gamma}$ is the mean of $\gamma$ and $\alpha$ is the target selecting ratio. This forces the selector to select the most important $\alpha$ tokens for each source input instead of keeping all of them. 
 
 In our VRS model, we can easily adjust the degree of controllability by limiting an upper bound of the conditional mutual information (CMI) $I(\beta,Y|X)$~\cite{zhao2018information}. Specifically, we can change our objective into:
\begin{equation}
\label{eq: vhard-obj}
\begin{split}
    \max_{\phi,\theta,\gamma}\mathbb{E}_{\beta \sim q_\phi}\log p_\theta(Y|X,\beta) \\- \lambda\abs{KL(q_\phi||\mathbf{B}(\gamma))-\epsilon)}
\end{split}
\end{equation}
$\lambda$ is a fixed lagrangian multiplier. Eq.~\ref{eq: vhard-obj} can be proved equal to maximum likelihood with the constraint $I(\beta,Y|X)=\epsilon$ given proper $\lambda$~\cite{pmlr-v80-alemi18a}. A higher $\epsilon$ indicates $\beta$ has more influences to $Y$ (higher controllability) while always safely selecting all tokens will lead $I(\beta,Y|X)=0$.\footnote{We also tried adding a coverage constraint to ensure the decoder covers all the selected tokens~\cite{wen2015semantically,wang2019toward}, but we find it brings no tangible help since a higher CMI can already discourage including redundant tokens into the selection.} It is preferred over Eq.~\ref{eq: hh-alpha} because (a) CMI directly considers the dependency between the selection and \emph{multiple}-possible text while limiting the ratio aims at finding the \emph{single} most salient parts for each source. (b) Unlike CMI, limiting the ratio is coarse. It considers only the total selected size and ignores its internal distribution.

\begin{algorithm}[tb]
   \caption{Variational Reinforce-Select (VRS)}
   \label{alg}
\begin{algorithmic}
   \STATE {\bfseries Parameters:} $\theta,\phi,\gamma$
   \STATE $pretrain \leftarrow$ TRUE
   \REPEAT
   \STATE Sample X,Y from the corpus;
   \STATE Encode X into $(h_1,h_2,\ldots,h_n)$;
   
   \IF{$pretrain$}
   \STATE Update $\phi$ with distant supervision;\\
   Update $\theta,\gamma$ by  $\nabla_{\theta,\gamma}$Eq.~\ref{eq: vhard-mask};
   \ELSE
   \STATE Update $\theta,\gamma,\phi$ by  $\nabla_{\theta,\gamma,\phi}$Eq.~\ref{eq: vhard-obj};
   \ENDIF
   \STATE $pretrain \leftarrow$ FALSE if Eq.~\ref{eq: vhard-mask} degrades
   \UNTIL{convergence and $pretrain$ is False}
\end{algorithmic}
\end{algorithm}

 In practice, we can set $\epsilon$ to adjust the degree of controllability we want. Later we will show it leads to a trade-off with performance. The final algorithm is detailed in Algorithm~\ref{alg}. To keep fairness, we trian RS and VRS with the same control variate and pre-training strategy.\footnote{The only extra parameter of VRS is $\phi$ which is a simple MLP structure. The actual training complexity is similar to RS because they both use the REINFORCE algorithm for gradient estimation.}

\section{Related Work}
Most content selection models train the selector with heuristic rules~\cite{hsu2018unified,li2018improving,yu2018operation,gehrmann2018bottom,yao2018plan,moryossef2019step}, which fail to fully capture the relation between selection and generation. \citet{mei2016talk,zhou2017selective,lin2018global,li2018improving} ``soft-select" word or sentence embeddings based on a gating function. The output score from the gate is a deterministic vector without any probabilistic variations, so controlling the selection to generate diverse text is impossible. Very
few works explicitly define a bernoulli distribution for the selector, then train with the REINFORCE algorithm~\cite{ling2017coarse,chen2018fast}, but the selection targets at a high recall regardless of the low precision, so the controllability over generated text is weak. \citet{fan2017controllable} control the generation by manually concatenating entity embeddings, while our model is much more flexible by explicitly defining the selection probability over all source tokens. 

Our work is closely related with learning discrete representations with variational inference~\cite{wen2017latent,van2017neural,kaiser2018fast,lawson2018learning}, where we treat content selection as the latent representation. 
Limiting the KL-term is a common technique to deal with the ``posterior collapse" problem ~\cite{kingma2016improved,yang2017improved,shen2018improving}. We adopt a similar approach and use it to further control the selecting strategy.
\section{Experiments}
For the experiments, we focus on comparing
(1) Bottom-up generation (Bo.Up.), (2) soft-select (SS), (3) Reinforce-select (RS) and (4) Variational-Reinforce-select (VRS) regarding their performance on content selection. SS and RS are trained with the selecting ratio constraint in Eq.~\ref{eq: hh-alpha}. For the SS model, we further add a regularization term to encourage the maximum value of $\gamma$ to be close to 1 as in \citet{mei2016talk}. We first briefly introduce the tasks and important setup, then present the evaluation results. 

\subsection{Tasks and Setup}
We test content-selection models on the headline and data-to-text generation task. Both tasks share the same framework with the only difference of source-side encoders. 

\textbf{Headline Generation}:
We use English Gigaword preprocessed by \citet{rush2015neural}, which pairs first sentences of news articles with their headlines. We keep most settings same as in \citet{zhou2017selective}, but use a vocabulary built by byte-pair-encoding~\cite{sennrich2016neural}. We find it speeds up training with superior performance.

\textbf{Data-to-Text Generation}:
We use the Wikibio dataset~\cite{lebret2016neural}.
The source is a Wikipedia infobox and the target is a one-sentence biography description. Most settings are the same as in \citet{liu2018table}, but we use a bi-LSTM encoder for better performance. 
 
\textbf{Heuristically extracted content}: This is used to train the selector for bottom up models and pre-train the RS and VRS model. For wikibio, we simply extract overlapped words between the source and target. In Gigaword, as the headline is more abstractive, we select the closest source word for each target word in the embedding space. Stop words and punctuations are prohibited from being selected.
\begin{table*}[t!]
\centering
\begin{tabular}{lrrr|rrrr}
\multirow{2}{*}{{Gigaword}} & \multicolumn{3}{c}{\small Oracle upper bound of} &{ \% unique }&{ \% unique}& {\% \small{Effect of}}&{\small Entropy of}\\
    & \small{ROUGE-1} & \small{ROUGE-2} &\small{ROUGE-L}{}& {Generation}& {Mask}&{Selector} & {Selector}\\
  \hline
  \small{Bo.Up.} &42.61 & 22.32 &38.37& 84.28& 95.87 & 87.91 & 0.360\\
  \small{SS} &33.15 & 14.63 &30.68& 82.06&\textbf{96.23} & 85.27 & \textbf{0.392}\\
  {\small RS} & 36.62 & 18.34 &34.60& 3.01& 6.23 & 48.31 & 0.018 \\
  \small{VRS} &\textbf{54.73} & \textbf{33.28} &\textbf{51.62}&\textbf{89.23}&92.51 & \textbf{96.45} & 0.288\\
  \hline
  {Wikibio} & \small{ROUGE-4} & \small{BLEU-4} &\small{NIST}& {Generation}& {Mask} & {Selector} & Selector \\
  \small{Bo.Up.} &47.28 & 49.95 &11.06& 31.57& 77.42 & 40.78 & 0.177 \\
  \small{SS} &41.73 & 43.94 &9.82& \textbf{63.09}& \textbf{89.42} & 70.55 & \textbf{0.355}\\
  {\small RS} & 44.07 & 46.89 &10.31& 4.55& 43.83 & 10.38&0.105 \\
  \small{VRS} &\textbf{52.41} & \textbf{55.03} &\textbf{11.89}& 57.62& 77.83 & \textbf{74.03} & 0.181\\
\end{tabular}
\caption{\label{tab: diverse}Diversity of content selection. The \% effect of selector is defined as the ratio of unique generation and mask, which reflects the rate that changing the selector will lead to corresponding changes of the generated text.
  }
\end{table*}

\textbf{Choice of $\alpha/\epsilon$}: As seen in Sec~\ref{sec: trade-off}, we need to set the hyperparameter $\alpha$ for RS/SS and $\epsilon$ for VRS. $\alpha$ corresponds to the selecting ratio. We set them as $\alpha=0.35$ for Wikibio and $0.25$ for Gigaword. The value is decided by running a human evaluation to get the empirical estimation. To keep comparison fairness, we tune $\epsilon$ to make VRS select similar amount of tokens with RS. The values we get are $\epsilon=0.15n$ for Wikibio and $\epsilon=0.25n$ for Gigaword. $n$ is the number of source tokens.~\footnote{$\epsilon$ corresponds to the KL divergence of the selection mask, which scales linearly with the number of source tokens, so we set it proportinally w.r.t. $n$.} 

\subsection{Results and Analysis}
Ideally we would expect the learned content selector to (1) have reasonable diversity so that text with various contents can be easily sampled, (2) properly control the contents described in the generated text and (3) not hurt performance. The following section will evaluate these three points in order.
\label{sec: results}

\textbf{Diversity}: We first look into the diversity of content selection learned by different models. For each test data, 50 selection masks are randomly sampled from the model's learned distribution. Greedy decoding is run to generate the text for each mask. We measure the entropy of the selector, proportion of unique selection masks and generated text in the 50 samples. We further define the ``effect" of the selector as the ratio of sampled unique text and mask. This indicates how often changing the selection mask will also lead to a change in the generated text. The results are averaged over all test data. Following \citet{rush2015neural} and \citet{lebret2016neural}, we measure the quality of generated text with ROUGE-1, 2, L F-score for Gigaword and ROUGE-4, BLEU-4, NIST for Wikibio. As there is only one reference text for each source, we report
an oracle upper bound of these scores by assuming an ``oracle" that can choose the best text among all the candidates~\cite{mao2014deep,wang2017diverse}. Namely, out of each 50 sampled text, we pick the one with the maximum metric score. The final metric score is evaluated on these ``oracle" picked samples. The intuition is that if the content selector is properly trained, at least one out of the 50 samples should describe similar contents with the reference text, the metric score between it and the reference text should be high. Table~\ref{tab: diverse} lists the results. We can have the following observations:
\begin{itemize}
    \item RS model completely fails to capture the content-level diversity. Its selector is largely deterministic, with a lowest entropy value among all models.
    In contrast, the selector from SS, VRS and Bo.Up. have reasonable diversity, with over $90\%$ and $75\%$ unique selection masks for Gigaword and Wikibio respectively.
    \item The selector from VRS has the strongest effect to the generator, especially on the Gigaword data where modifying the content selection changes the corresponding text in more than 95\% of the cases. RS has the lowest effect value, which indicates that even with the selecting ratio constraint, its generator still ignores the selection mask to a large extent.
    \item The oracle metric score of VRS is much higher than the other two. This is beneficial when people want to apply the model to generate a few candidate text then hand-pick the suitable one. VRS has more potential than the other three to contain the expected text. SS performs worst. The gap between the soft approximation and the real distribution, as mentioned before, indeed results in a large drop of performance.
\end{itemize}
In short, compared with others, the content selector of VRS is \emph{(1) diverse, (2) has stronger effect on the text generation and (3) with a larger potential of producing an expected text}.

\textbf{Controllability}: We have shown the content selector of VRS is diverse and has strong effect on the text generation. This section aims at examining whether such effect is desirable, i.e., whether the selector is able to properly control the contents described in the text. We measure it based on the self-bleu metric and a human evaluation.

 The self-bleu metric measures the controllability by evaluating the ``intra-selection" similarity of generated text. Intuitively, by fixing the selection mask, multiple text sampled from the decoder are expected to describe the same contents and thereby should be highly similar to each other. The decoder should only model surface-level diversity without further modifying the selected contents. With this in mind, for each test data, we randomly sample a selection mask from the selector's distribution, then fix the mask and run the decoder to sample 10 different text. The self-BLEU-1 score~\cite{zhu2018texygen} on the sampled text is reported, which is the average BLEU score between each text pair. A higher self-BLEU score indicates the sampled text are more similar with each other. The results are shown in Table~\ref{tab: self-bleu}. We can see generations from VRS have a clearly higher intra-selection similarity. SS performs even worse than RS, despite having a high effect score in Table~\ref{tab: diverse}. The selector from SS affects the generation in an undesirable way, which also explain why SS has a lowest oracle metric score though with a high score on content diversity and effect.
 
\begin{table}[!ht]
\centering
\begin{tabular}{@{}llllll@{}}
\toprule
Method & Bo.Up. &SS & RS & VRS \\ \midrule
Gigaword & 46.58 & 37.20 & 48.13 & \textbf{61.14}\\
Wikibio & 38.30 & 13.92 & 25.99 & \textbf{43.81} \\
\bottomrule
\end{tabular}
\caption{Self-Bleu score by fixing selection mask. Higher means better controllability of content selection}
\label{tab: self-bleu}
\end{table}

\begin{table}[t]
\centering
\begin{tabular}{@{}cccc@{}}
\toprule
Method & \small{Fluency} & \small{intra-consistency} & \small{inter-diversity}\\ \midrule
\small{Reference}& 0.96 &-  &- \\
\small{Enc-Dec} & 0.83 &  -&-  \\\hline
\small{Bo.Up.} & 0.46 & 0.48 & 0.61 \\
\small{SS} & 0.27 & 0.41 & 0.54\\
\small{RS} & \textbf{0.78} & 0.39 & 0.47\\
\small{VRS} &0.74 & \textbf{0.72} & \textbf{0.87}\\
\bottomrule
\end{tabular}
\caption{Human evaluation on fluency, intra-consistency and inter-diversity of content selection on DUC 2004.}
\label{tab: human-eval}
\end{table}

 We further run a human evaluation to measure the text-content consistency among different models. 100 source text are randomly sampled from the human-written DUC 2004 data for task 1\&2~\cite{over2007duc}. Bo.Up, SS, RS and VRS are applied to generate the target text by first sampling a selection mask, then run beam search decoding with beam size 10. We are interested in seeing (1) if multiple generations from the same selection mask are paraphrases to each other (intra-consistent) and (2) if generations from different selection masks do differ in the content they described (inter-diverse). The results in Table~\ref{tab: human-eval} show that VRS significantly outperforms the other two in both intra-consistency and inter-diversity. RS has the lowest score on both because the selector has very weak effects on the generation as measured in the last section. Bo.Up and SS lay between them. Overall VRS is able to \emph{maintain the highest content-text consistency} among them.
 
 \begin{table}[ht]
\centering
\begin{tabular}{@{}lllll@{}}
\toprule
\small{Method} & \small{R-1} & \small{R-2} & \small{R-L} &\small{\%Word} \\ \midrule
\small{\citet{zhou2017selective}} & 36.15 & 17.54 & 33.63 & 100
\\
\small{Enc-Dec} & 35.92 & 17.43 & 33.42 & 100\\\hline
\small{SS} & 20.35 & 4.78 & 16.53 &24.82\\
\small{Bo.Up} & 28.17 & 10.32 & 26.68 &24.54\\
\small{RS} & 35.45 & 16.38 & 32.71 &25.12\\
\small{VRS($\epsilon=0$)-pri} &\textbf{36.42} & \textbf{17.81} & \textbf{33.86} &78.63\\
\small{VRS($\epsilon=0.25$)-pri} & 34.26 & 15.11 & 31.69 &24.36\\\hline\hline
\small{VRS($\epsilon=0$)-post} &37.14 & 18.03 & 34.26 &78.66\\
\small{VRS($\epsilon=0.25$)-post} &\textbf{56.72} & \textbf{33.24} & \textbf{51.88} &24.53\\
\bottomrule
\end{tabular}
\caption{Gigaword best-select results. Larger $\epsilon$ leads to more controllable selector with a bit degrade of performance. (-post) means selecting from the posterior $q_\phi(\beta|X,Y)$, (-pri) is from the prior $\mathbf{B}(\gamma_i)$.}
\label{tab: giga-accuracy}
\end{table}

\begin{table}[t]
\centering
\begin{tabular}{@{}lllll@{}}
\toprule
\small{Method} & \small{R-4} & \small{B-4} & \small{NIST} &\small{\%Word} \\ \midrule
\small{\citet{liu2018table}} & 41.65 & 44.71 &  & 100\\
\small{Enc-Dec} & 42.07 & 44.80 & 9.82 & 100\\\hline
\small{SS} & 5.10 & 5.73 & 0.24 &35.12\\
\small{Bo.Up} & 8.07 & 9.52 & 0.42 &38.79\\
\small{RS} & 42.64 & 45.08 & 10.01 &34.53\\
\small{VRS($\epsilon=0$)-pri} &\textbf{43.01} & \textbf{46.01} & \textbf{10.24} &84.56\\
\small{VRS($\epsilon=0.15$)-pri} & 42.13 & 44.51 & 9.84 &34.04\\\hline\hline
\small{VRS($\epsilon=0$)-post} &43.84 & 46.60 & 10.27 &85.34\\
\small{VRS($\epsilon=0.15$)-post} &\textbf{49.68} & \textbf{52.26} & \textbf{11.48} &34.57\\
\bottomrule
\end{tabular}
\caption{Wikibio best-select results.}
\label{tab: wiki-accuracy}
\end{table}

\textbf{Performance $\&$ Trade-off}: To see if the selector affects performance, we also ask human annotators to judge the text fluency. The fluency score is computed as the average number of text being judged as fluent. We include generations from the standard Enc-Dec model. Table~\ref{tab: human-eval} shows the best fluency is achieved for Enc-Dec. Imposing a content selector always affects the fluency a bit. The main reason is that when the controllability is strong, the change of selection will directly affect the text realization so that a tiny error of content selection might lead to unrealistic text. If the selector is not perfectly trained, the fluency will inevitably be influenced. When the controllability is weaker, like in RS, the fluency is more stable because it will not be affected much by the selection mask. For SS and Bo.Up, the drop of fluency is significant because of the gap of soft approximation and the independent training procedure. In general, VRS does properly decouple content selection from the enc-dec architecture, with only tiny degrade on the fluency.

Table~\ref{tab: giga-accuracy}/\ref{tab: wiki-accuracy} further measure the metric scores on Gigaword/Wikibio by decoding text from the best selection mask based on the selector's distribution (set $\beta_i=1$ if $\mathbf{B}(\gamma_i)>0.5$ and $0$ otherwise). We include results from VRS model with $\epsilon=0$, which puts no constraint on the mutual information. We further report the score by generating the best selection mask from the learned posterior distribution $q_\phi(\beta|X,Y)$ for VRS model. Two current SOTA results from \citet{zhou2017selective} and \citet{liu2018table} and the proportion of selected source words for each model are also included. We have the following observations:
\begin{itemize}
    \item As the value of $\epsilon$ decreases, the performance of VRS improves, but the selector loses more controllability because the model tends to over-select contents (over $75\%$ source words selected). The text-content consistency will become low.
    \item Increasing $\epsilon$ sacrifices a bit performance, but still comparable with SOTA. Especially on Wikibio where the performance drop is minor. The reason should be that Wikibio is relatively easier to predict the selection but Gigaword has more uncertainty.
    \item Increasing $\epsilon$ improves the accuracy of the posterior selection. This would be useful when we want to perform posterior inference for some source-target pair.
    \item Setting $\epsilon=0$ can actually outperform SOTA seq2seq which keeps all tokens, suggesting it is still beneficial to use the VRS model even if we do not care about the controllability.
\end{itemize}

\begin{figure}[!ht]
\centering
\centerline{\includegraphics[width=\columnwidth]{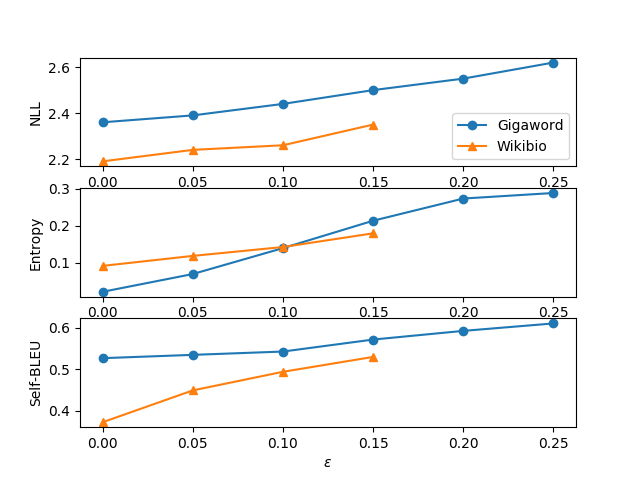}}
\caption{Negative log likelihood (NLL), selection entropy and self-BLEU as $\epsilon$ changes. NLL and self-bleu on Wikibio are added by 1 for better visualization. Lower NLL suggests higher performance. Higher entropy/self-BLEU means higher diversity/controllability.}
\label{fig: epsilon}
\end{figure}

Figure~\ref{fig: epsilon} visualizes how changing the value of $\epsilon$ affects the negative log likelihood (NLL), entropy of the selector and self-bleu score, which roughly correlates with performance, diversity and controllability. NLL is evaluated based on the lower bound in Eq~\ref{eq: vhard-mask}~\cite{sohn2015learning}.
We can see as $\epsilon$ increases, the performance decreases gradually but the content selection gains more diversity and controllability. In practice we can tune the $\epsilon$ value to achieve a trade-off. 

\begin{figure}[!ht]
\begin{boxedminipage}{\columnwidth}
\small
\textbf{Source:} \textcolor{gred}{indian prime minister} p.v. narasimha rao 's \textcolor{gred}{promise} of more autonomy for troubled \textcolor{gred}{kashmir} and his plea for early state elections has \textcolor{gred}{sparked} a \textcolor{gred}{violent reaction} from provincial moslem and opposition parties .
\newline 
\textbf{Samples from SS:}\newline
\textbf{t1:}  indian indian calls for end to violence in kashmir .\newline
\textbf{t2:} indian pm calls for end to violence in afghanistan .\newline
\textbf{t3:} indian pm calls for boycott of pakistan 's ruling party .\newline
\textbf{Samples from Bo.Up:}\newline
\textbf{t1:}  india promises more autonomous more autonomy .\newline
\textbf{t2:} indian pm promises autonomy for kashmir autonomy .\newline
\textbf{t3:} indian pm 's promise sparks violent reaction .\newline
\textbf{Samples from RS:}\newline
\textbf{t1:}  indian pm 's kashmir promises sparks violent reaction.\newline
\textbf{t2:} indian pm 's promise sparks violent reaction .\newline
\textbf{t3:} kashmir parties blast pm 's promise .\newline
\textbf{Samples from VRS:}
\newline\textbf{t1:}  indian pm 's promise on kashmir sparks uproar .\newline
\textbf{t2:} indian pm 's promise on kashmir sparks protests .\newline
\textbf{t3:} indian pm 's promise for kashmir sparks controversy .
\newline
\rule{\columnwidth}{0.4pt}
\textbf{Source:}  \textcolor{gred}{factory orders} for manufactured goods \textcolor{gred}{rose \#.\# percent in september} , the commerce department said here thursday . \newline
\textbf{Samples from SS:}\newline
\textbf{t1:}  u.s. consumer confidence down in january in january.\newline
\textbf{t2:} u.s. wholesale prices up \#.\# percent in october .\newline
\textbf{t3:} u.s. jobless rate rises to \#.\# percent in march .
\newline 
\textbf{Samples from Bo.Up.:}\newline
\textbf{t1:}  september u.s. factory orders up \#.\# percent .\newline
\textbf{t2:} september u.s. factory orders increase .\newline
\textbf{t3:} factory orders up in september .
\newline
\textbf{Samples from RS:}\newline
\textbf{t1:}  u.s. factory orders up \#.\# percent in september .\newline
\textbf{t2:} factory orders for manufactured goods rise .\newline
\textbf{t3:} factory orders up in september from the year .
\newline
\textbf{Samples from VRS:}\newline
\textbf{t1:}  september factory orders up \#.\# percent .\newline
\textbf{t2:} september factory orders rise \#.\# percent .\newline
\textbf{t3:} september factory orders increase \#.\# pct .
\end{boxedminipage}
\caption{Text generation examples from Gigaword. \textcolor{gred}{Highlighted} words are selected. t1-3 are sampled from the decoder based on the selected content. Generations from VRS are more faithful to selected contents.}
\label{fig: gigaword-example}
\end{figure}
\textbf{Generation Example}: Figure~\ref{fig: gigaword-example} shows some examples from Gigaword. As can be seen, decodings from the VRS model are largely consistent with each other, in most cases only replacing one or two words with corresponding synonyms. Samples are able to faithfully convey all selected contents. In contrast, generations from SS. Bo.Up. and RS are unpreditable, differing in both selected contents and also the way of saying. SS and Bo.Up also suffer more from the text disfluency. The generations from them are largely uncertain.

\section{Conclusion}
In this paper, we tackle the unaddressed problem of controllable content selection in text generation. We propose a general framework based on variational inference that can be potentiall applied to arbitrary tasks. On both the headline generation and data-to-text tasks, our model outperforms state-of-the-art models regarding the diversity and controllability of content selection. We further introduce an effective way to achieve a performance/controllability trade-off, which can be easily tuned to meet specific requirement. 
\section*{Acknowledgments}
We thank anonymous reviewers for valuable comments, thank Aditya Mogadala, Shun Kiyono, Thomas Mclachlan and other members of the LIAT team at RIKEN AIP for useful discussions. Xiaoyu Shen is supported by IMPRS-CS fellowship. The work of J. Suzuki was partly supported by JSPS KAKENHI Grant Number JP19104418 and AIRPF Grant Number 30AI036-8. This work is also partially funded by DFG collaborative
research center SFB 1102.


\bibliography{emnlp-ijcnlp-2019}

\begin{thebibliography}{64}
\expandafter\ifx\csname natexlab\endcsname\relax\def\natexlab#1{#1}\fi

\bibitem[{Alemi et~al.(2018)Alemi, Poole, Fischer, Dillon, Saurous, and
  Murphy}]{pmlr-v80-alemi18a}
Alexander Alemi, Ben Poole, Ian Fischer, Joshua Dillon, Rif~A. Saurous, and
  Kevin Murphy. 2018.
\newblock Fixing a broken {ELBO}.
\newblock In \emph{Proceedings of the 35th International Conference on Machine
  Learning}, volume~80 of \emph{Proceedings of Machine Learning Research},
  pages 159--168, StockholmsmÃ€ssan, Stockholm Sweden. PMLR.

\bibitem[{Bahdanau et~al.(2015)Bahdanau, Cho, and Bengio}]{bahdanau2015neural}
Dzmitry Bahdanau, Kyunghyun Cho, and Yoshua Bengio. 2015.
\newblock Neural machine translation by jointly learning to align and
  translate.
\newblock In \emph{International Conference on Learning Representations}.

\bibitem[{Burda et~al.(2016)Burda, Grosse, and
  Salakhutdinov}]{burda2015importance}
Yuri Burda, Roger Grosse, and Ruslan Salakhutdinov. 2016.
\newblock Importance weighted autoencoders.
\newblock In \emph{International Conference on Learning Representations}.

\bibitem[{Chen and Bansal(2018)}]{chen2018fast}
Yen-Chun Chen and Mohit Bansal. 2018.
\newblock Fast abstractive summarization with reinforce-selected sentence
  rewriting.
\newblock In \emph{Proceedings of the 56th Annual Meeting of the Association
  for Computational Linguistics (Volume 1: Long Papers)}, pages 675--686.
  Association for Computational Linguistics.

\bibitem[{Cornia et~al.(2018)Cornia, Baraldi, and Cucchiara}]{cornia2018show}
Marcella Cornia, Lorenzo Baraldi, and Rita Cucchiara. 2018.
\newblock Show, control and tell: A framework for generating controllable and
  grounded captions.
\newblock \emph{arXiv preprint arXiv:1811.10652}.

\bibitem[{Deng et~al.(2018)Deng, Kim, Chiu, Guo, and Rush}]{deng2018latent}
Yuntian Deng, Yoon Kim, Justin Chiu, Demi Guo, and Alexander~M Rush. 2018.
\newblock Latent alignment and variational attention.
\newblock \emph{Nips}.

\bibitem[{Devlin et~al.(2019)Devlin, Chang, Lee, and
  Toutanova}]{devlin2018bert}
Jacob Devlin, Ming-Wei Chang, Kenton Lee, and Kristina Toutanova. 2019.
\newblock Bert: Pre-training of deep bidirectional transformers for language
  understanding.
\newblock \emph{NAACL}.

\bibitem[{Fan et~al.(2018)Fan, Grangier, and Auli}]{fan2017controllable}
Angela Fan, David Grangier, and Michael Auli. 2018.
\newblock Controllable abstractive summarization.
\newblock In \emph{Proceedings of the 2nd Workshop on Neural Machine
  Translation and Generation}, pages 45--54. Association for Computational
  Linguistics.

\bibitem[{Gehrmann et~al.(2018)Gehrmann, Deng, and Rush}]{gehrmann2018bottom}
Sebastian Gehrmann, Yuntian Deng, and Alexander~M Rush. 2018.
\newblock Bottom-up abstractive summarization.
\newblock \emph{EMNLP}.

\bibitem[{Glynn(1990)}]{glynn1990likelilood}
Peter~W. Glynn. 1990.
\newblock Likelihood ratio gradient estimation for stochastic systems.
\newblock \emph{Commun. ACM}, 33(10):75--84.

\bibitem[{Grathwohl et~al.(2018)Grathwohl, Choi, Wu, Roeder, and
  Duvenaud}]{grathwohl2018backpropagation}
Will Grathwohl, Dami Choi, Yuhuai Wu, Geoff Roeder, and David Duvenaud. 2018.
\newblock Backpropagation through the void: Optimizing control variates for
  black-box gradient estimation.
\newblock In \emph{International Conference on Learning Representations}.

\bibitem[{Haury et~al.(2011)Haury, Gestraud, and Vert}]{haury2011influence}
Anne-Claire Haury, Pierre Gestraud, and Jean-Philippe Vert. 2011.
\newblock The influence of feature selection methods on accuracy, stability and
  interpretability of molecular signatures.
\newblock \emph{PloS one}, 6(12):e28210.

\bibitem[{Hsu et~al.(2018)Hsu, Lin, Lee, Min, Tang, and Sun}]{hsu2018unified}
Wan-Ting Hsu, Chieh-Kai Lin, Ming-Ying Lee, Kerui Min, Jing Tang, and Min Sun.
  2018.
\newblock A unified model for extractive and abstractive summarization using
  inconsistency loss.
\newblock In \emph{Proceedings of the 56th Annual Meeting of the Association
  for Computational Linguistics (Volume 1: Long Papers)}, pages 132--141.
  Association for Computational Linguistics.

\bibitem[{Jackson(1998)}]{jackson1998introduction}
Peter Jackson. 1998.
\newblock \emph{Introduction to expert systems}.
\newblock Addison-Wesley Longman Publishing Co., Inc.

\bibitem[{Kaiser et~al.(2018)Kaiser, Bengio, Roy, Vaswani, Parmar, Uszkoreit,
  and Shazeer}]{kaiser2018fast}
Lukasz Kaiser, Samy Bengio, Aurko Roy, Ashish Vaswani, Niki Parmar, Jakob
  Uszkoreit, and Noam Shazeer. 2018.
\newblock Fast decoding in sequence models using discrete latent variables.
\newblock In \emph{Proceedings of the 35th International Conference on Machine
  Learning}, volume~80 of \emph{Proceedings of Machine Learning Research},
  pages 2390--2399, StockholmsmÃ€ssan, Stockholm Sweden. PMLR.

\bibitem[{Ke et~al.(2018)Ke, {\.Z}o{\l}na, Sordoni, Lin, Trischler, Bengio,
  Pineau, Charlin, and Pal}]{pmlr-v80-ke18a}
Nan~Rosemary Ke, Konrad {\.Z}o{\l}na, Alessandro Sordoni, Zhouhan Lin, Adam
  Trischler, Yoshua Bengio, Joelle Pineau, Laurent Charlin, and Christopher
  Pal. 2018.
\newblock Focused hierarchical {RNN}s for conditional sequence processing.
\newblock In \emph{Proceedings of the 35th International Conference on Machine
  Learning}, volume~80 of \emph{Proceedings of Machine Learning Research},
  pages 2554--2563, StockholmsmÃ€ssan, Stockholm Sweden. PMLR.

\bibitem[{Kingma et~al.(2016)Kingma, Salimans, Jozefowicz, Chen, Sutskever, and
  Welling}]{kingma2016improved}
Diederik~P Kingma, Tim Salimans, Rafal Jozefowicz, Xi~Chen, Ilya Sutskever, and
  Max Welling. 2016.
\newblock Improved variational inference with inverse autoregressive flow.
\newblock In \emph{Advances in Neural Information Processing Systems}, pages
  4743--4751.

\bibitem[{Kingma and Welling(2014)}]{kingma2013auto}
Diederik~P Kingma and Max Welling. 2014.
\newblock Auto-encoding variational bayes.
\newblock In \emph{International Conference on Learning Representations}.

\bibitem[{Lawson et~al.(2018)Lawson, Chiu, Tucker, Raffel, Swersky, and
  Jaitly}]{lawson2018learning}
Dieterich Lawson, Chung-Cheng Chiu, George Tucker, Colin Raffel, Kevin Swersky,
  and Navdeep Jaitly. 2018.
\newblock Learning hard alignments with variational inference.
\newblock In \emph{2018 IEEE International Conference on Acoustics, Speech and
  Signal Processing (ICASSP)}, pages 5799--5803. IEEE.

\bibitem[{Lebret et~al.(2016)Lebret, Grangier, and Auli}]{lebret2016neural}
R{\'e}mi Lebret, David Grangier, and Michael Auli. 2016.
\newblock Neural text generation from structured data with application to the
  biography domain.
\newblock In \emph{Proceedings of the 2016 Conference on Empirical Methods in
  Natural Language Processing}, pages 1203--1213. Association for Computational
  Linguistics.

\bibitem[{Lei et~al.(2016)Lei, Barzilay, and Jaakkola}]{lei2016rationalizing}
Tao Lei, Regina Barzilay, and Tommi Jaakkola. 2016.
\newblock Rationalizing neural predictions.
\newblock In \emph{Proceedings of the 2016 Conference on Empirical Methods in
  Natural Language Processing}, pages 107--117.

\bibitem[{Li et~al.(2018)Li, Xiao, Lyu, and Wang}]{li2018improving}
Wei Li, Xinyan Xiao, Yajuan Lyu, and Yuanzhuo Wang. 2018.
\newblock Improving neural abstractive document summarization with explicit
  information selection modeling.
\newblock \emph{EMNLP}.

\bibitem[{Lin et~al.(2018)Lin, SUN, Ma, and Su}]{lin2018global}
Junyang Lin, Xu~SUN, Shuming Ma, and Qi~Su. 2018.
\newblock Global encoding for abstractive summarization.
\newblock In \emph{Proceedings of the 56th Annual Meeting of the Association
  for Computational Linguistics (Volume 2: Short Papers)}, pages 163--169.
  Association for Computational Linguistics.

\bibitem[{Ling and Rush(2017)}]{ling2017coarse}
Jeffrey Ling and Alexander Rush. 2017.
\newblock Coarse-to-fine attention models for document summarization.
\newblock In \emph{Proceedings of the Workshop on New Frontiers in
  Summarization}, pages 33--42.

\bibitem[{Liu et~al.(2018)Liu, Wang, Sha, Chang, and Sui}]{liu2018table}
Tianyu Liu, Kexiang Wang, Lei Sha, Baobao Chang, and Zhifang Sui. 2018.
\newblock Building end-to-end dialogue systems using generative hierarchical
  neural network models.
\newblock In \emph{Proceedings of the Thirty-Second AAAI Conference on
  Artificial Intelligence}, pages 4881--4888. AAAI Press.

\bibitem[{Luong et~al.(2015)Luong, Pham, and Manning}]{luong2015effective}
Thang Luong, Hieu Pham, and Christopher~D. Manning. 2015.
\newblock Effective approaches to attention-based neural machine translation.
\newblock In \emph{Proceedings of the 2015 Conference on Empirical Methods in
  Natural Language Processing}, pages 1412--1421. Association for Computational
  Linguistics.

\bibitem[{Ma et~al.(2017)Ma, Gao, Hu, Yu, Deng, and Hovy}]{ma2017dropout}
Xuezhe Ma, Yingkai Gao, Zhiting Hu, Yaoliang Yu, Yuntian Deng, and Eduard Hovy.
  2017.
\newblock Dropout with expectation-linear regularization.
\newblock In \emph{International Conference on Learning Representations}.

\bibitem[{Mao et~al.(2015)Mao, Xu, Yang, Wang, Huang, and Yuille}]{mao2014deep}
Junhua Mao, Wei Xu, Yi~Yang, Jiang Wang, Zhiheng Huang, and Alan Yuille. 2015.
\newblock Deep captioning with multimodal recurrent neural networks (m-rnn).
\newblock \emph{ICLR}.

\bibitem[{Mei et~al.(2016)Mei, UChicago, Bansal, and Walter}]{mei2016talk}
Hongyuan Mei, TTI UChicago, Mohit Bansal, and Matthew~R Walter. 2016.
\newblock What to talk about and how? selective generation using lstms with
  coarse-to-fine alignment.
\newblock In \emph{Proceedings of NAACL-HLT}, pages 720--730.

\bibitem[{Mnih and Gregor(2014)}]{mnih2014neural}
Andriy Mnih and Karol Gregor. 2014.
\newblock Neural variational inference and learning in belief networks.
\newblock In \emph{Proceedings of the 31st International Conference on
  International Conference on Machine Learning - Volume 32}, ICML'14, pages
  II--1791--II--1799. JMLR.org.

\bibitem[{Mnih and Rezende(2016)}]{mnih2016variational}
Andriy Mnih and Danilo~J. Rezende. 2016.
\newblock Variational inference for monte carlo objectives.
\newblock In \emph{Proceedings of the 33rd International Conference on
  International Conference on Machine Learning - Volume 48}, ICML'16, pages
  2188--2196. JMLR.org.

\bibitem[{Moryossef et~al.(2019)Moryossef, Goldberg, and
  Dagan}]{moryossef2019step}
Amit Moryossef, Yoav Goldberg, and Ido Dagan. 2019.
\newblock Step-by-step: Separating planning from realization in neural
  data-to-text generation.
\newblock \emph{NAACL}.

\bibitem[{Nallapati et~al.(2017)Nallapati, Zhai, and
  Zhou}]{nallapati2017summarunner}
Ramesh Nallapati, Feifei Zhai, and Bowen Zhou. 2017.
\newblock Summarunner: A recurrent neural network based sequence model for
  extractive summarization of documents.
\newblock In \emph{AAAI}.

\bibitem[{Nema et~al.(2017)Nema, Khapra, Laha, and
  Ravindran}]{nema2017diversity}
Preksha Nema, Mitesh~M Khapra, Anirban Laha, and Balaraman Ravindran. 2017.
\newblock Diversity driven attention model for query-based abstractive
  summarization.
\newblock In \emph{Proceedings of the 55th Annual Meeting of the Association
  for Computational Linguistics (Volume 1: Long Papers)}, volume~1, pages
  1063--1072.

\bibitem[{van~den Oord et~al.(2017)van~den Oord, Vinyals, and
  kavukcuoglu}]{van2017neural}
Aaron van~den Oord, Oriol Vinyals, and koray kavukcuoglu. 2017.
\newblock Neural discrete representation learning.
\newblock In I.~Guyon, U.~V. Luxburg, S.~Bengio, H.~Wallach, R.~Fergus,
  S.~Vishwanathan, and R.~Garnett, editors, \emph{Advances in Neural
  Information Processing Systems 30}, pages 6306--6315. Curran Associates, Inc.

\bibitem[{Over et~al.(2007)Over, Dang, and Harman}]{over2007duc}
Paul Over, Hoa Dang, and Donna Harman. 2007.
\newblock Duc in context.
\newblock \emph{Information Processing \& Management}, 43(6):1506--1520.

\bibitem[{Pennington et~al.(2014)Pennington, Socher, and
  Manning}]{pennington2014glove}
Jeffrey Pennington, Richard Socher, and Christopher Manning. 2014.
\newblock Glove: Global vectors for word representation.
\newblock In \emph{Proceedings of the 2014 conference on empirical methods in
  natural language processing (EMNLP)}, pages 1532--1543.

\bibitem[{Prabhumoye et~al.(2019)Prabhumoye, Quirk, and
  Galley}]{prabhumoye2019towards}
Shrimai Prabhumoye, Chris Quirk, and Michel Galley. 2019.
\newblock Towards content transfer through grounded text generation.
\newblock \emph{NAACL}.

\bibitem[{Qin et~al.(2018)Qin, Yao, Wang, Wang, and Lin}]{qin2018learning}
Guanghui Qin, Jin-Ge Yao, Xuening Wang, Jinpeng Wang, and Chin-Yew Lin. 2018.
\newblock Learning latent semantic annotations for grounding natural language
  to structured data.
\newblock In \emph{Proceedings of the 2018 Conference on Empirical Methods in
  Natural Language Processing}, pages 3761--3771.

\bibitem[{Raiko et~al.(2015)Raiko, Berglund, Alain, and
  Dinh}]{raiko2014techniques}
Tapani Raiko, Mathias Berglund, Guillaume Alain, and Laurent Dinh. 2015.
\newblock Techniques for learning binary stochastic feedforward neural
  networks.
\newblock \emph{ICLR}.

\bibitem[{Reiter and Dale(2000)}]{reiter2000building}
Ehud Reiter and Robert Dale. 2000.
\newblock \emph{Building natural language generation systems}.
\newblock Cambridge university press.

\bibitem[{Rush et~al.(2015)Rush, Chopra, and Weston}]{rush2015neural}
Alexander~M. Rush, Sumit Chopra, and Jason Weston. 2015.
\newblock A neural attention model for abstractive sentence summarization.
\newblock In \emph{Proceedings of the 2015 Conference on Empirical Methods in
  Natural Language Processing}, pages 379--389. Association for Computational
  Linguistics.

\bibitem[{Sennrich et~al.(2016)Sennrich, Haddow, and
  Birch}]{sennrich2016neural}
Rico Sennrich, Barry Haddow, and Alexandra Birch. 2016.
\newblock Neural machine translation of rare words with subword units.
\newblock In \emph{Proceedings of the 54th Annual Meeting of the Association
  for Computational Linguistics (Volume 1: Long Papers)}, volume~1, pages
  1715--1725.

\bibitem[{Shen et~al.(2018{\natexlab{a}})Shen, Zhou, Long, Jiang, Wang, and
  Zhang}]{shen2018reinforced}
Tao Shen, Tianyi Zhou, Guodong Long, Jing Jiang, Sen Wang, and Chengqi Zhang.
  2018{\natexlab{a}}.
\newblock Reinforced self-attention network: a hybrid of hard and soft
  attention for sequence modeling.
\newblock In \emph{Proceedings of the Twenty-Seventh International Joint
  Conference on Artificial Intelligence, {IJCAI-18}}, pages 4345--4352.
  International Joint Conferences on Artificial Intelligence Organization.

\bibitem[{Shen et~al.(2018{\natexlab{b}})Shen, Su, Niu, and
  Demberg}]{shen2018improving}
Xiaoyu Shen, Hui Su, Shuzi Niu, and Vera Demberg. 2018{\natexlab{b}}.
\newblock Improving variational encoder-decoders in dialogue generation.
\newblock In \emph{Thirty-Second AAAI Conference on Artificial Intelligence}.

\bibitem[{Sohn et~al.(2015)Sohn, Lee, and Yan}]{sohn2015learning}
Kihyuk Sohn, Honglak Lee, and Xinchen Yan. 2015.
\newblock Learning structured output representation using deep conditional
  generative models.
\newblock In \emph{Advances in neural information processing systems}, pages
  3483--3491.

\bibitem[{Stent et~al.(2004)Stent, Prasad, and Walker}]{stent2004trainable}
Amanda Stent, Rashmi Prasad, and Marilyn Walker. 2004.
\newblock Trainable sentence planning for complex information presentation in
  spoken dialog systems.
\newblock In \emph{Proceedings of the 42nd annual meeting on association for
  computational linguistics}, page~79. Association for Computational
  Linguistics.

\bibitem[{Sutskever et~al.(2014)Sutskever, Vinyals, and
  Le}]{sutskever2014sequence}
Ilya Sutskever, Oriol Vinyals, and Quoc~V Le. 2014.
\newblock Sequence to sequence learning with neural networks.
\newblock In Z.~Ghahramani, M.~Welling, C.~Cortes, N.~D. Lawrence, and K.~Q.
  Weinberger, editors, \emph{Advances in Neural Information Processing Systems
  27}, pages 3104--3112. Curran Associates, Inc.

\bibitem[{Tucker et~al.(2019)Tucker, Lawson, Gu, and
  Maddison}]{tucker2019doubly}
George Tucker, Dieterich Lawson, Shixiang Gu, and Chris~J Maddison. 2019.
\newblock Doubly reparameterized gradient estimators for monte carlo
  objectives.
\newblock In \emph{International Conference on Learning Representations}.

\bibitem[{Tucker et~al.(2017)Tucker, Mnih, Maddison, Lawson, and
  Sohl-Dickstein}]{tucker2017rebar}
George Tucker, Andriy Mnih, Chris~J Maddison, John Lawson, and Jascha
  Sohl-Dickstein. 2017.
\newblock Rebar: Low-variance, unbiased gradient estimates for discrete latent
  variable models.
\newblock In I.~Guyon, U.~V. Luxburg, S.~Bengio, H.~Wallach, R.~Fergus,
  S.~Vishwanathan, and R.~Garnett, editors, \emph{Advances in Neural
  Information Processing Systems 30}, pages 2627--2636. Curran Associates, Inc.

\bibitem[{Wang et~al.(2017)Wang, Schwing, and Lazebnik}]{wang2017diverse}
Liwei Wang, Alexander Schwing, and Svetlana Lazebnik. 2017.
\newblock Diverse and accurate image description using a variational
  auto-encoder with an additive gaussian encoding space.
\newblock In \emph{Advances in Neural Information Processing Systems}, pages
  5756--5766.

\bibitem[{Wang et~al.(2019)Wang, Hu, Yang, Shi, Xu, and Xing}]{wang2019toward}
Wentao Wang, Zhiting Hu, Zichao Yang, Haoran Shi, Frank Xu, and Eric Xing.
  2019.
\newblock Toward unsupervised text content manipulation.
\newblock \emph{arXiv preprint arXiv:1901.09501}.

\bibitem[{Weaver and Tao(2001)}]{weaver2001optimal}
Lex Weaver and Nigel Tao. 2001.
\newblock The optimal reward baseline for gradient-based reinforcement
  learning.
\newblock In \emph{Proceedings of the Seventeenth Conference on Uncertainty in
  Artificial Intelligence}, UAI'01, pages 538--545, San Francisco, CA, USA.
  Morgan Kaufmann Publishers Inc.

\bibitem[{Wen et~al.(2015)Wen, Gasic, Mrk{\v{s}}i{\'c}, Su, Vandyke, and
  Young}]{wen2015semantically}
Tsung-Hsien Wen, Milica Gasic, Nikola Mrk{\v{s}}i{\'c}, Pei-Hao Su, David
  Vandyke, and Steve Young. 2015.
\newblock Semantically conditioned lstm-based natural language generation for
  spoken dialogue systems.
\newblock In \emph{Proceedings of the 2015 Conference on Empirical Methods in
  Natural Language Processing}, pages 1711--1721.

\bibitem[{Wen et~al.(2017)Wen, Miao, Blunsom, and Young}]{wen2017latent}
Tsung-Hsien Wen, Yishu Miao, Phil Blunsom, and Steve Young. 2017.
\newblock Latent intention dialogue models.
\newblock In \emph{Proceedings of the 34th International Conference on Machine
  Learning}, volume~70 of \emph{Proceedings of Machine Learning Research},
  pages 3732--3741, International Convention Centre, Sydney, Australia. PMLR.

\bibitem[{Williams(1992)}]{williams1992simple}
Ronald~J. Williams. 1992.
\newblock Simple statistical gradient-following algorithms for connectionist
  reinforcement learning.
\newblock \emph{Machine Learning}, 8(3):229--256.

\bibitem[{Wiseman et~al.(2018)Wiseman, Shieber, and Rush}]{wiseman2018learning}
Sam Wiseman, Stuart~M Shieber, and Alexander~M Rush. 2018.
\newblock Learning neural templates for text generation.
\newblock \emph{EMNLP}.

\bibitem[{Xu et~al.(2015)Xu, Ba, Kiros, Cho, Courville, Salakhudinov, Zemel,
  and Bengio}]{xu2015show}
Kelvin Xu, Jimmy Ba, Ryan Kiros, Kyunghyun Cho, Aaron Courville, Ruslan
  Salakhudinov, Rich Zemel, and Yoshua Bengio. 2015.
\newblock Show, attend and tell: Neural image caption generation with visual
  attention.
\newblock In \emph{International conference on machine learning}, pages
  2048--2057.

\bibitem[{Yang et~al.(2017)Yang, Hu, Salakhutdinov, and
  Berg-Kirkpatrick}]{yang2017improved}
Zichao Yang, Zhiting Hu, Ruslan Salakhutdinov, and Taylor Berg-Kirkpatrick.
  2017.
\newblock Improved variational autoencoders for text modeling using dilated
  convolutions.
\newblock \emph{ICML}.

\bibitem[{Yao et~al.(2019)Yao, Peng, Ralph, Knight, Zhao, and
  Yan}]{yao2018plan}
Lili Yao, Nanyun Peng, Weischedel Ralph, Kevin Knight, Dongyan Zhao, and Rui
  Yan. 2019.
\newblock Plan-and-write: Towards better automatic storytelling.
\newblock \emph{AAAI}.

\bibitem[{Yu et~al.(2018)Yu, Zhang, Huang, and Zhu}]{yu2018operation}
Naitong Yu, Jie Zhang, Minlie Huang, and Xiaoyan Zhu. 2018.
\newblock An operation network for abstractive sentence compression.
\newblock In \emph{Proceedings of the 27th International Conference on
  Computational Linguistics}, pages 1065--1076. Association for Computational
  Linguistics.

\bibitem[{Zhao et~al.(2018)Zhao, Song, and Ermon}]{zhao2018information}
Shengjia Zhao, Jiaming Song, and Stefano Ermon. 2018.
\newblock The information autoencoding family: A lagrangian perspective on
  latent variable generative models.
\newblock \emph{UAI}.

\bibitem[{Zhou et~al.(2017)Zhou, Yang, Wei, and Zhou}]{zhou2017selective}
Qingyu Zhou, Nan Yang, Furu Wei, and Ming Zhou. 2017.
\newblock Selective encoding for abstractive sentence summarization.
\newblock In \emph{Proceedings of the 55th Annual Meeting of the Association
  for Computational Linguistics (Volume 1: Long Papers)}, pages 1095--1104.
  Association for Computational Linguistics.

\bibitem[{Zhu et~al.(2018)Zhu, Lu, Zheng, Guo, Zhang, Wang, and
  Yu}]{zhu2018texygen}
Yaoming Zhu, Sidi Lu, Lei Zheng, Jiaxian Guo, Weinan Zhang, Jun Wang, and Yong
  Yu. 2018.
\newblock Texygen: a benchmarking platform for text generation models.
\newblock In \emph{The 41st International ACM SIGIR Conference on Research \&
  Development in Information Retrieval}, pages 1097--1100. ACM.

\end{thebibliography}
\bibliographystyle{acl_natbib}

\clearpage
\appendix

\section{Performance/Controllability trade-off}
\label{app: trade-off}
The trade-off between performance and interpretability has been a long-standing problem in feature selection~\cite{jackson1998introduction,haury2011influence}. The trade-off exists because it is usually very difficult to accurately find the exact features needed to make the prediction. Safely keeping more features will almost always lead to better performance. Some models do succeed in achieving superior performance by selecting only a subset of the input. However, they mostly still target at the recall of the selection~\cite{hsu2018unified,chen2018fast,shen2018reinforced}, i.e., to select all possible content that might help predict the target. The final selected contents reduce some most useful information from the source, but they still contain many redundant contents (same like our VRS-($\epsilon=0$) as in Table~\ref{tab: wiki-accuracy} and \ref{tab: giga-accuracy}). This makes them unsuitable for controllable content selection. In text generation, a recent work from \citet{moryossef2019step} shows they could control the contents by integrating a symbolic selector into the neural network. However, their selector is tailored by some rules only for the RDF triples. Moreover, even based on their fine-tuned selector, the fluency they observe is still slightly worse than a standard seq2seq.

We assume the content selector is the major bottle if we want a model that can achieve controllability without sacrificing the performance. We can clearly observe in Table~\ref{tab: wiki-accuracy} that the performance drop in Wikibio is marginal compared with Gigaword. The reason should be that the selection on Wikibio is much easier than Gigaword. The biography of a person almost always follow some simple patterns, like name, birthday and profession, but for news headlines, it can contain information with various focuses. In our two tasks, due to the independence assumption we made on $\beta_i$ and the model capacity limit, the content selector cannot fully fit the true selecting distribution, so the trade-off is necessary. Improving the selector with SOTA sequence labelling models like Bert~\cite{devlin2018bert} would be worth trying.

There are also other ways to improve. For example, we could learn a ranker to help us choose the best contents~\cite{stent2004trainable}. Or we could manually define some matching rules to help rank the selection~\cite{cornia2018show}. In Table~\ref{tab: diverse}, we show the VRS model achieves very high metric scores based on an oracle ranker, so learning a ranker should be able to improve the performance straightforwardly.

\begin{figure}[ht]
\begin{boxedminipage}{\columnwidth}
\small
\textbf{Source:} The \textcolor{red}{sri lankan} government on Wednesday announced the \textcolor{red}{closure} of government \textcolor{red}{schools} with immediate effect \textcolor{red}{as} a \textcolor{red}{military} campain against tamil separatists \textcolor{red}{escalated} in the north of the country. \newline
\textbf{Reference:} sri lanka closes schools as war escalates .\newline
\textbf{b1:}  sri lanka shuts schools as war escalates .\newline
\textbf{b2:} sri lanka closes schools as violence escalates .\newline
\textbf{b3:} sri lanka shuts schools as fighting escalates .\newline
\textbf{b4:} sri lanka closes schools as offensive expands .\newline
\textbf{b5:} sri lanka closes schools as war continues .
\end{boxedminipage}
\caption{Posterior inference example. \textcolor{red}{Highlighted} words are selected contents according to the posterior distribution $q_{\phi}(\beta|X,Y)$. b1-b5 are decoded by fixing the selected contents.}
\label{fig: post-example}
\end{figure}

\section{Example from Wikibio}
 \begin{figure*}
\centering
\centerline{\includegraphics[width=2\columnwidth]{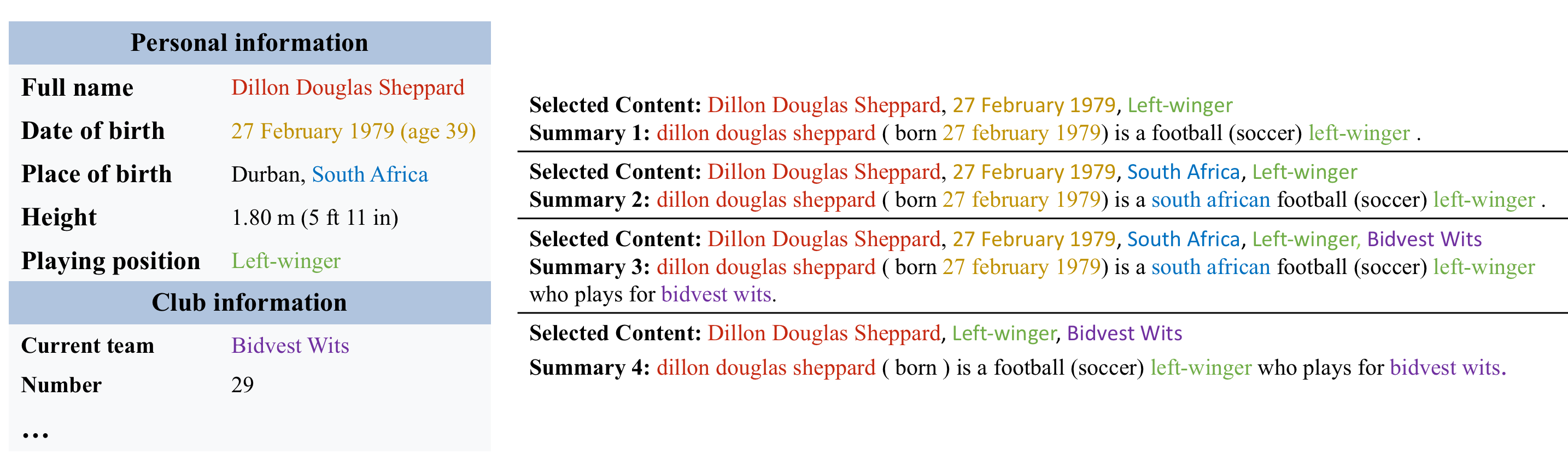}}
\caption{Example of content selection in Wikibio. Summary 4 in unnatural because we force to remove the birthday content which the selector assigns very high probability to select.}
\label{fig: wikibio-example}
\end{figure*}
To see how we can manually control the content selection, Figure~\ref{fig: wikibio-example} shows an example from Wikibio, the model is mostly able to form a proper sentence covering all selected information. If the selector assigns very high probability to select some content and we force to remove it, the resulting text could be unnatual (as in summary 4 in Figure~\ref{fig: wikibio-example} because the model has seen very few text without containing the birthday information in the training corpus). However, thanks to the diversity of the content selector as shown in the previous section, it is able to handle most combinatorial patterns of content selection.

\section{Posterior inference}
\label{app: post}
Figure~\ref{fig: post-example} further provides an example of how we can perform posterior inference given a provided text. Our model is able to infer which source contents are covered in the given summary. With the inferred selection, we can sample multiple paraphrases describing the same contents. As seen in Table~\ref{tab: wiki-accuracy} and \ref{tab: giga-accuracy}, the metric scores are remarkably high when decoding from the posterior inferred selections (last three rows), suggesting the posterior distribution is well trained. The posterior inference part could be beneficial for other tasks like content transfer among text~\cite{wang2019toward,prabhumoye2019towards}. The described source contents can be first predicted with the posterior inference, then transferred to a new text.
 
\clearpage
\onecolumn
\begin{table}[h]
\begin{tabularx}{\textwidth}{l|XX}
          & Gigaword & Wikibio \\
\small Vocabulary & \small Built with byte-pair segmentation with size 30k&  \small Built by keeping the most frequent 20k tokens\\
\small Word Embedding & \multicolumn{2}{c}{\small Size 300. Initialized with Glove~\cite{pennington2014glove}. OOVs are randomly initialized from a normal distribution}\\
\small Inputs & \small Sequence of source word embeddings&  \small Sequence of concatenation of source table field, value, position and reverse position embeddings~\cite{lebret2016neural}\\
\hline
\small Source Encoder &  \small Single-layer Bi-LSTM with hidden size 512 & \small Single-layer Bi-LSTM with hidden size 500\\
\small Target Decoder & \small Single-layer LSTM with hidden size 512& \small Single-layer LSTM with hidden size 500\\\small Drop out rate & \multicolumn{2}{c}{\small 0.3 for both encoder and decoder}\\\small Decoding Method & \small Beam search with beam size 5& \small Greedy decoding. UNK words are replaced with the most attended source token as in \citet{liu2018table}\\
\hline
\small Mini-batch size & \small 256& \small 128\\\hline
\small Optimizer & \multicolumn{2}{c}{\small Adam, $\beta_1=0.9,\beta_2=0.999,\epsilon=10^{-8},\text{weight decay}=1.2\times 10^{-6}$, gradient clipping in [-5,5]}\\
\small Initial Learning Rate & \multicolumn{2}{c}{\small 0.0005} \\\hline
\small Prior Selector &\multicolumn{2}{c}{\small $\mathbf{B}(\gamma_i)=\sigma(\textmd{MLP}(h_i))$. MLP is a multi-layer perceptron. $h_i$ comes from the encoder hidden state}\\
\small Posterior Selector &\multicolumn{2}{c}{\small $q_\phi(\beta_i|X,Y)=\sigma(\textmd{MLP}([h_i \circ e(y)]))$. $\circ$ means concatenation}
\end{tabularx}
\caption{\label{tab: setting}Detailed settings of our experiment. $e(y)$ in the posterior selector is an encoded representation of the ground-truth text. We use a bi-LSTM encoder. The last hidden state is treated as the representation for the text.}
\end{table}
\end{document}